\documentclass[11pt,a4paper]{article}
\usepackage[hyperref]{naaclhlt2019}
\usepackage{times}
\usepackage{latexsym}
\usepackage{graphicx}

\usepackage{url}

\newcommand\size[1]{|\mathbf{#1}|}
\newcommand\ALy[0]{\mathrm{AL}_{\size{y}}}
\newcommand\AL[0]{\mathrm{AL}}
\newcommand\AFL[0]{\mathrm{DAL}}

\aclfinalcopy 

\title{Thinking Slow about Latency Evaluation \\for Simultaneous Machine Translation}

\author{
    Colin Cherry \and George Foster\\
    Google\\
    {\tt $\{$colincherry, fosterg$\}$@google.com}}

\date{}

\begin{document}
\maketitle

\begin{abstract}
Simultaneous machine translation attempts to translate a source sentence
before it is finished being spoken,
with applications to translation of spoken language for live streaming and 
conversation.
Since simultaneous systems trade quality to reduce latency,
having an effective and interpretable latency metric is crucial.
We introduce a variant of the recently proposed Average Lagging ($\AL$) metric, 
which we call Differentiable Average Lagging ($\AFL$). 
It distinguishes itself by being differentiable and internally consistent to its underlying 
mathematical model.
\end{abstract}
  
\section{Introduction}
Simultaneous machine translation begins translating the source
sentence before it is finished, sacrificing some translation quality
in order to reduce \textbf{latency}:
the amount of time the target language consumer spends waiting for
their translation while the source language speaker is speaking.
The trade-off between latency and quality is central to simultaneous
MT, making the accurate measurement of latency crucial.
However, the community has yet to settle on a standard latency metric,
especially for the intrinsic scenario, where we are working on
source sentences with no timing information, and
delay must be estimated based on the rate at which the MT system
consumes source tokens.
The underlying assumption of these intrinsic metrics is that the only
appreciable source of latency in a simultaneous translation occurs when
the system opts to wait to read the next source token.

The Average Lagging ($\AL$) latency metric has been recently proposed by
\newcite{Ma2018} to measure the average rate by which an MT system
lags behind an ideal translator that is completely simultaneous with
the source language producer.
While this metric is a big step forward in terms of its
interpretability and its careful handling of differences in source and
target sentence lengths, its current formulation is not
differentiable.
Furthermore, we argue that AL is built on top of
inconsistent assumptions; in particular,
it is inconsistent in its treatment of how long it takes the
MT system to write a target token.
%
We
show that by clearly stating and reasoning about
assumptions, we can develop a metric that maintains the spirit and
positive properties of AL while also being differentiable.
We dub this new metric Differentiable Average Lagging ($\AFL$).

\section{Background}

We are concerned with calculating latency for a previously-written
source sentence, without further source-speaker timing information, as
is necessary when evaluating on standard MT training, development or
test sets.
In this scenario, all timing information is derived from the rate
at which source tokens are consumed by the MT system.

We adopt the notation of \newcite{Ma2018}, which in turn adopts a
formalism popularized by \newcite{Grissom2014}, where the simultaneous
MT system consists of an \textbf{agent} that begins with an empty
source sentence, 
and must select between \textbf{read} actions that
reveal source tokens for use in translation, and \textbf{write}
actions that produce target tokens, both operating one token at a time
and from left to right.
Let $\mathbf{x}$ and $\mathbf{y}$ be source and target sequences,
and let $t$, $1 \leq t \leq \size{y}$, index the target sequence.
Our primary data structure for calculating latency will be $g(t)$, a
function that gives the number of source tokens read by the agent
before writing target token $t$.
Standard (non-simultaneous) MT systems have $\forall t$:
$g(t)=\size{x}$,
as they read the entire source sequence before writing any target tokens.

\subsection{Previous latency metrics}

Before the advent of neural machine translation, work on simultaneous
MT tended to report either the latency of end-to-end systems in
milliseconds~\cite{Bangalore2012,RangarajanSridhar2013},
%
or with method-specific metrics that are only loosely correlated with
latency, such as the number of target tokens per source segment for
segmentation-based approaches~\cite{RangarajanSridhar2013,Oda2014}.
An interesting exception is \newcite{Grissom2014}, who opt instead to
measure latency and translation quality with a single metric, Latency
BLEU, which averages BLEU scores (with brevity penalty) calculated on
the (potentially empty) partial translations available after each
source token is read.

Alongside the first strategies for neural simultaneous MT,
\newcite{Cho16} introduced the Average Proportion (AP) metric, which
averages the absolute source delay incurred by each target token:
\begin{equation}
  \mathrm{AP}=
  \frac{1}{\size{x}\phantom{\cdot}\size{y}}\sum_{t=1}^{\size{y}}g(t)
\end{equation}
This metric has some nice properties, such as being bound between 0
and 1, but it also has some issues.
\newcite{Ma2018} observe that their \mbox{wait-$k$} system%
\footnote{%
  A simultaneous MT system which reads $k$ source tokens, and then
  proceeds with a write-1-read-1 pattern until the entre source
  sequence has been read: \mbox{$g(t)=\min(t+k-1, \size{x})$}.}
with a fixed $k=1$ incurs different AP values as sequence
length $\size{x}=\size{y}$ ranges from 2 ($\mathrm{AP}=0.75$)
to $\infty$ ($\mathrm{AP}=0.5$).
Knowing that a very-low-latency wait-1 system incurs at best an
$\mathit{AP}$ of 0.5 also implies that much of the metric's dynamic
range is wasted; in fact, 
\newcite{Alinejad2018} report that AP is not sufficiently sensitive to
detect their improvements to a reinforcement learning approach to
simultaneous MT.

\newcite{Gu2017} use AP, and also introduce the position-wise latency
metric Consecutive Wait (CW), which measures the number of consecutive
reads between writes:
\begin{equation}
  \mathrm{CW}(t) = \left\{
    \begin{array}{cl}
      g(t)-g(t-1) & t > 1 \\
      g(t) & t = 1 \\
    \end{array} \right.
\end{equation}
Though CW has not officially been extended to a metric of sentence-level
latency, we note that \newcite{Alinejad2018} report average-CW in response
to the lack of sensitivity in AP.

Recently, \newcite{Ma2018} introduced Average Lagging ($\AL$),
which measures the average rate by which the MT system lags behind an
ideal wait-0 translator:
\begin{equation}
  \label{eq:AL}
  \mathrm{AL} = \frac{1}{\tau}\sum_{t=1}^{\tau} g(t)-\frac{t-1}{\gamma}
\end{equation}
where $\tau$ is the earliest timestep where the MT system has consumed
the entire source sequence:
\[
\tau = \mathrm{argmin}_{t} \: [g(t)=\size{x}]
\]
and $\gamma = \size{y}/\size{x}$ accounts for the source and
target having different sequence lengths.
This metric has the nice property that when $\size{x}=\size{y}$,
a wait-$k$ system will achieve an AL of $k$.
Furthermore, when $\size{y} > \size{x}$,
$\gamma$ forces a wait-$k$ system to 
\textbf{catch up}, by occasionally writing multiple target tokens consecutively,
in order to achieve an AL of $k$.

\section{Differentiable Average Lagging}

\subsection{Problems with Average Lagging}

Our problems with Average Lagging begin with its inability to be optimized.
The $\mathrm{argmin}$ operation used to calculate $\tau$ is not differentiable.
%
%
Since our problem stems from $\tau$, we will now carefully consider
why $\tau$ is there.

\newcite{Ma2018} do not discuss $\tau$'s purpose, but we can infer it
from a few examples by comparing to a simpler version of $\AL$ where
$\tau=\size{y}$:
\begin{equation}
  \label{eq:ALy}
  \ALy = \frac{1}{\size{y}}\sum_{t=1}^{\size{y}}
  g(t)-\frac{t-1}{\gamma}
\end{equation}
Why not use $\ALy$?
Because it does not fulfill the desiderata of having average lagging
be equal to $k$ for a wait-$k$ system.
Table~\ref{tab:exEqualLengths} illustrates this for $k=1$ ($\ALy=1$) and
$k=3$ ($\ALy=2.25$).
\begin{table*}[t]
  \begin{center}
  \begin{tabular}{lcccc|cc}
    \multicolumn{7}{c}{$k=1$} \\
    \hline
    $t$       & 1 & 2 & 3 & 4 & &  \\
    $g(t)$    & 1 & 2 & 3 & 4 & $\tau=4$ &   \\
    $\ell(t)$ & 1 & 1 & 1 & 1 & $\AL=1$  & $\ALy=1$ \\
    
  \end{tabular}
  \begin{tabular}{||lcccc|cc}
    \multicolumn{7}{c}{$k=3$}  \\
    \hline
    $t$       & 1 & 2 & 3 & 4 &  &  \\
    $g(t)$    & 3 & 4 & 4 & 4 & $\tau=2$   &  \\
    $\ell(t)$ & 3 & 3 & 2 & 1 & $\AL=3$  & $\ALy=2.25$ \\
  \end{tabular}
  \end{center}
  \caption{Time-indexed lag
    $\ell(t)=g(t)-\frac{t-1}{\gamma}$ when $\size{x}=\size{y}=4$ for
    wait-$k$ systems with $k=1$ and $k=3$. \label{tab:exEqualLengths}}
\end{table*}
Looking at the time-indexed lags $\ell$ for the $k=3$ scenario, the
problem with $\ALy$ becomes clear:
each position where $g(t)=\size{x}$ beyond the first has its lag
reduced by 1, pulling the average lag below $k$.
$\AL$'s solution to this problem is equally clear: by stopping at
$\tau$, we omit the problematic indexes from the average.
We argue that $\tau$ is a patch on top of a more fundamental problem.

Why are indexes $t>\tau$ problematic? Because they allow the wait-$k$
system to exploit an assumption in our metric: that time advances only
as we read source tokens.
After all source tokens have been read, all remaining target tokens
appear instantaneously, reducing the lag for later tokens.
By asserting that timesteps $t>\tau$ do not contribute to $\AL$, we
are implicitly asserting that the assumption of instant or free writes 
does not make sense:
the system lagged behind the source speaker while they were speaking;
it should continue to lag by the same amount after they finish
speaking.

The argument in favour of the simpler $\ALy$ is that in a text-based
system, we can effectively write instantaneously; from a human's
perspective, all characters can appear on the screen at once.
The argument in favour of $\AL$'s $\tau$ is that we will often be in a
speech-to-speech scenario where it takes time to speak each token, and
even in the pure text-output scenario, it still takes our human reader
time to read each token.

Let's accept this latter argument and assert that $\AL$ is preferable
to $\ALy$.\footnote{While also acknowledging that this is an
  assumption to enable a mathematical model to drive an artificial
  metric.}
%
Therefore, there should be a cost to emitting target tokens when
$t>\tau$.
One way to implement such a cost is to stop averaging at $\tau$.
Now the question becomes: why should we only incur this cost while
$t>\tau$?
As currently specified, $\AL$ can continue to benefit from free
writes so long as $t\leq\tau$; that is,
%
while the source sentence is still being read. 
%
After this point,
writes incur some poorly specified
cost, equivalent to ignoring the remaining target tokens.
We argue that this inconsistent and undesirable.

\subsection{The consequences of free writes}

$\AL$ relies on the special-casing of $t>\tau$ to maintain
its intuitive results.
This special case is not easily abused by the deterministic wait-$k$
strategies that $\AL$ has been used to evaluate thus far,
but future adaptive schedules may exploit it.
Compare for example two systems in the scenario where $\size{x}=\size{y}=5$: 
\begin{enumerate}
\item a standard wait-4 system: read 4, write 1, read 1, write 4.
\item a similar system that delays the final read: read 4, write 4,  read 1, write 1.
\end{enumerate}
The two systems differ only in when they read the final token.
The corresponding $g$ and $l$ values are shown in Table~\ref{tab:al_adversarial}.
Note that they have very similar $g$ values: identical for $t=1$ and 5, and differing only by 1 for $t=2$, 3 and 4.
\begin{table*}[t]
  \begin{center}
  \begin{tabular}{lcccccc}
    $t$       & 1 & 2 & 3 & 4 & 5 & $\AL$\\
    $g(t)$    & 4 & 5 & 5 & 5 & 5 & \\
    $\ell(t)$ & 4 & 4 & - & - & - & 4\\
  \end{tabular}
  \begin{tabular}{||lcccccc}
    $t$       & 1 & 2 & 3 & 4 & 5 & $\AL$ \\
    $g(t)$    & 4 & 4 & 4 & 4 & 5 & \\
    $\ell(t)$ & 4 & 3 & 2 & 1 & 1 & 2.2 \\
  \end{tabular}
  \end{center}
  \caption{Time-indexed lag
    $\ell(t)=g(t)-\frac{t-1}{\gamma}$ when $\size{x}=\size{y}=5$ for a
    standard wait-4 system (left) and for an antagonistic system that delays its final read (right). The $\AL$ of the former is 4, while the $\AL$ of the latter is 2.2.\label{tab:al_adversarial}}
\end{table*}
However the latter system has been engineered exploit $\AL$'s structure,
and by delaying its final read, it has reduced its $\AL$ from 4 to 2.2.

\subsection{Writing with costs}

Now we introduce Differentiable Average Lagging ($\AFL$) which alters $\AL$ to
maintain the desiderata:
\begin{enumerate}
\item a wait-$k$ system should incur a lag of $k$, and
\item lag should account for sentence lengths when $\size{y} \neq \size{x}$,
\end{enumerate}
while also consistently accounting for the cost of writing target
tokens.
Along the way, we will eliminate $\tau$, creating a metric that is
differentiable.

A key insight in our design is that the problem with $\AL$ begins with
$g(t)$, which measures delay only in terms of number of source tokens
read.
Let $d$ be the time-cost (also measured in number of source tokens)
for writing a target token.
%
%
We construct a $g'$ that wraps $g$ in a model that accounts for 
target-writing costs:
\begin{equation}
  g_d'(t) = \left\{
  \begin{array}{ll}
    g(t) & t=1\\
    \max\big[g(t), g_d'(t-1)+d\big] & t>1
  \end{array}\right.
\end{equation}
$g_d'(t)$ tracks how much source time has passed immediately
before writing the target token $t$, mirroring the semantics of $g(t)$.
The second term of the $\max$ represents a baseline minimum time:
the amount of time that passed immediately before the previous target 
token, plus the cost of writing that token.
The first term, which represents reading $g(t)$ source tokens,
will not add any more delay to $g'$, unless it exceeds the second term;
that is,
some source tokens are available to be read ``for free''
because that much source time has already passed.

This new $g'$ gives us one half of our metric.
The other half is the ideal timing for each position, which is represented by
$\frac{t-1}{\gamma}$ in $\AL$.
We can derive our ideal timing by reasoning about an MT system without 
latency.
Conceptually, the simplest latency-free translator is prescient; it never
reads the source and therefore never delays.
For this ideal system, ${g}(t) = 0$ for all $t$,
meaning ${g}_d'(t)=(t-1)d$.
Using this as our ideal timing term gives us the parameterized metric:
\begin{equation}
  \label{eq:AFLd}
  \AFL_d = \frac{1}{\size{y}}\sum_{t=1}^{\size{y}}
  g_d'(t)-(t-1)d
\end{equation}

We could leave the cost of target writes $d$ as a hyper-parameter to be
set depending on the scenario, but we recommend
$d=\frac{1}{\gamma}=\frac{\size{x}}{\size{y}}$ for three reasons.
First, it maintains consistency with $\AL$, creating a final metric
that is quite similar:
\begin{equation}
  \label{eq:AFL}
  \AFL = \frac{1}{\size{y}}\sum_{t=1}^{\size{y}}
  g'(t)-\frac{t-1}{\gamma}
\end{equation}
where $g'(t)=g'_{\frac{1}{\gamma}}(t)$.
Second, our ideal latency-free translator would finish speaking after
$\size{y}d=\size{y}\frac{\size{x}}{\size{y}}=\size{x}$ source units,
perfectly in sync with our source speaker.\footnote{Just like on Star
  Trek!}
%
%
Finally, $d=\size{x}/\size{y}$ ensures that $d<1$ when $\size{y}>\size{x}$,
which is necessary to encourage the system to catch up by writing several
tokens after a single read.

Note that we have eliminated $\tau$ and all $\mathrm{argmin}$
operations from $\AL$.
The recursion in $g'$ is differentiable, and can be implemented efficiently in
computation-graph-based programming languages using techniques similar 
to those used to enable recurrent neural networks.

\subsection{A non-recurrent formulation of $g'$}

For a concept so simple as delay with consistent writing costs,
our $g'$ solution might seem unnecessarily complex.
Unfortunately, a dependency on previous timesteps is necessary in order
to maintain a memory of previously incurred delays, but there is an
equivalent non-recurrent version, which expands the $\max$ to cover all
earlier timesteps.
\begin{equation}
    g_d'(t) = (t-1)d + \max_{1\leq i \leq t} \left[g(i)-(i-1)d\right]
    \label{eq:nonrecurrent}
\end{equation}
The equivalence of these two formulations can be proved by induction.
The non-recurrent formulation makes a few properties of $g'$ clear.
The lower-bound $g'(t) \geq (t-1)d$, which we leveraged earlier when building our ideal timing model, now stands out.
%
This formulation also shows how we incur further delay on top of this base according to whatever previous read (modeled by $g(i)$)
has gone the most over budget, where the budget is represented by $(i-1)d$.
Once the budget has been exceeded, the $\max$ ensures that $g'$ irrevocably incurs this delay for all future timesteps; 
delays can only increase as time passes.

Also note that when we plug Equation~\ref{eq:nonrecurrent} into Equation~\ref{eq:AFLd}, the $(t-1)d$ terms cancel,
making it clear that $\AFL$ only pays attention to the most expensive read thus far at each timestep.

\section{Discussion}

\subsection{Example scenarios}
\begin{table*}[t]
  \begin{center}
  \begin{tabular}{lcccc|c}
    \multicolumn{6}{c}{$k=1$} \\
    $\frac{t-1}{\gamma}$ & 0 & 1 & 2 & 3 & $\AFL$ \\
    $g(t)$    & 1 & 2 & 3 & 4     \\
    $g'(t)$    & 1 & 2 & 3 & 4     \\
    $\ell'(t)$ & 1 & 1 & 1 & 1 & 1  \\
    
  \end{tabular}
  \begin{tabular}{||lcccc|c}
    \multicolumn{6}{||c}{$k=3$}  \\
    $\frac{t-1}{\gamma}$ & 0 & 1 & 2 & 3 & $\AFL$ \\
    $g(t)$     & 3 & 4 & 4 & 4 &  \\
    $g'(t)$    & 3 & 4 & 5 & 6 &  \\
    $\ell'(t)$ & 3 & 3 & 3 & 3 & 3 \\
  \end{tabular}
  \end{center}
  \caption{$\AFL$ time-indexed lag
    $\ell'(t)=g'(t)-\frac{t-1}{\gamma}$ when $\size{x}=\size{y}=4$, and therefore $\gamma=1$; for wait-$k$ systems with $k=1$ and $k=3$. \label{tab:exAFL_EqualLengths}}
\end{table*}

We present a few illustrative examples of $\AFL$'s time-indexed lag
values.
Table~\ref{tab:exAFL_EqualLengths} returns to our equal sentence length, wait-1 and wait-3 scenarios.
As one can see, our use of $g'$ allows us to get the desired $\AFL$=$k$ for
both $k=1$ and $k=3$, while summing over all target indexes.

\begin{table*}[t]
  \begin{center}
  \begin{tabular}{lcccccc}
    $t$       & 1 & 2 & 3 & 4 & 5 & $\AFL$\\
    $g(t)$    & 4 & 5 & 5 & 5 & 5 & \\
    $g'(t)$    & 4 & 5 & 6 & 7 & 8 & \\
    $\ell'(t)$ & 4 & 4 & 4 & 4 & 4 & 4\\
  \end{tabular}
  \begin{tabular}{||lcccccc}
    $t$       & 1 & 2 & 3 & 4 & 5 & $\AFL$ \\
    $g(t)$    & 4 & 4 & 4 & 4 & 5 & \\
    $g'(t)$    & 4 & 5 & 6 & 7 & 8 & \\
    $\ell'(t)$ & 4 & 4 & 4 & 4 & 4 & 4 \\
  \end{tabular}
  \end{center}
  \caption{$\AFL$ Time-indexed lag
    $\ell'(t)=g'(t)-\frac{t-1}{\gamma}$ when $\size{x}=\size{y}=5$ for a
    standard wait-4 system (left) and for an antagonistic system that delays its final read (right). The $\AFL$ of both systems is 4.\label{tab:dal_adversarial}}
\end{table*}

\begin{figure}[t]
  \begin{center}
    \includegraphics[width=0.45\textwidth, clip=true, trim=10 380 400 10]{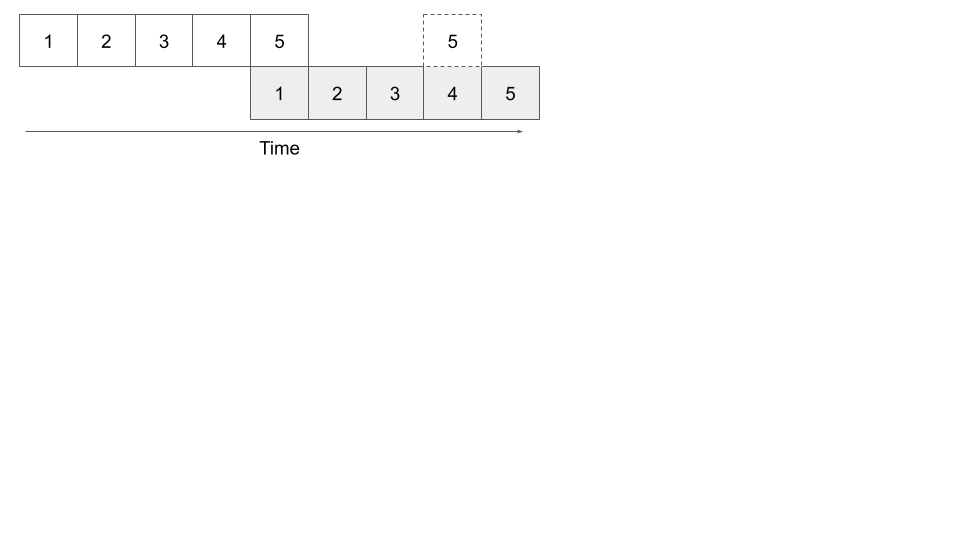}
    \caption{Illustration of the $\AFL$ score for the antagonistic example in Table~\ref{tab:dal_adversarial}. The top is the source as it streams in,
    the bottom is the streaming output, and the dotted box indicates where source token 5 is read.}
    \label{fig:dal_adversarial}
  \end{center}
\end{figure}

Next, we return to our motivating, antagonistic delayed-read system in Table~\ref{tab:dal_adversarial}.
One can see that both the wait-4 system and the antagonistic system receive $\AFL$ values of 4.
In fact, both systems receive identical $g'$ values.
This highlights a crucial, but counter-intuitive property of $\AFL$: after waiting for 4 source tokens, 
the $5^{\mathit{th}}$ source token is counted against the system at $t=2$, regardless of when the system reads it.
To help make this more intuitive, Figure~\ref{fig:dal_adversarial} provides an illustration of how $g'$ positions each target token along a timeline.

\begin{table*}[t]
  \begin{center}
  \begin{tabular}{cl|cccccc|c|cccccc|c}
    & &\multicolumn{7}{|c|}{$k=1$ without catch-up} &
    \multicolumn{7}{c}{$k=1$ with catch-up} \\ \hline
    & $\frac{t-1}{\gamma}$ & 0 & 0.5 & 1 & 1.5 & 2 & 2.5 & Metric
    &  0 & 0.5 & 1 & 1.5 & 2 & 2.5 & Metric \\ 
    \hline
    $\AL$ & $g(t)$    & 1 & 2 & 3 & 3 & 3 & 3 &
    & 1 & 1 & 2 & 2 & 3 & 3 & \\
    & $\ell(t)$ & 1 & 1.5 & 2 & -- & -- & -- & 1.5 
    & 1 & 0.5 & 1 & 0.5 & 1 & -- & 0.8 \\ \hline
    $\AFL$ & $g'(t)$    & 1 & 2 & 3 & 3.5 & 4 & 4.5 &    
    & 1 & 1.5 & 2 & 2.5 & 3 & 3.5 & \\
    & $\ell'(t)$ & 1 & 1.5 & 2 & 2 & 2 & 2 & 1.75  
    & 1 & 1 & 1 & 1 & 1 & 1 & 1\\
  \end{tabular}
  \end{center}
  \caption{Examples of $\AL$ and $\AFL$ time-indexed lags when $\size{x}=3$
  and $\size{y}=6$, and therefore $\gamma=2$, using wait-1 strategies with
  and without catch-up. \label{tab:exAFL_UnequalLengths}}
\end{table*}
Table~\ref{tab:exAFL_UnequalLengths} presents an unequal sentence-length
scenario for both $\AL$ and $\AFL$, using both a vanilla wait-1 system
and a wait-1 system that catches up at a rate proportional to the difference
in sentence lengths (writing 2 tokens for each read).
Looking at the no catch-up scenario, we see that the main difference between
the two metrics is that $\AFL$ sums over the entire sequence, resulting
in a slightly higher average.
Turning to the catch-up scenario on the right,
the $\AL$ rows demonstrate an instance of that metric's
use of free writes that take 0 time.
These allow even-indexed tokens to incur a time-indexed lag of only 0.5, 
resulting in an average of $0.8 < k=1$.
This is not the case for $\AFL$, which maintains a consistent lag of 1 
over all timesteps.
Note that $\AFL$'s $d=0.5$ rewards the catch-up strategy of writing 
two tokens for each read, without resorting to free writes.

\begin{table}[t]
  \begin{center}
  \begin{tabular}{cl|ccc|c}
    & $\frac{t-1}{\gamma}$ & 0 & 2 & 4 & Metric \\ \hline
    $\AL$ & $g(t)$      & 1 & 2 &  3 & \\
          & $\ell(t)$   & 1 & 0 & -1 & 0 \\ \hline
    $\AFL$ & $g'(t)$    & 1 & 3 &  5 & \\    
           & $\ell'(t)$ & 1 & 1 &  1 & 1 \\  
    \end{tabular}
  \end{center}
  \caption{Examples of $\AL$ and $\AFL$ time-indexed lags when $\size{x}=6$
  and $\size{y}=3$, and therefore $\gamma=\frac{1}{2}$, using a wait-1 strategy 
  without catch-up. \label{tab:exAFL_UnequalLengths_2}}
\end{table}

Finally, in Table~\ref{tab:exAFL_UnequalLengths_2} we examine unequal lengths from
the other direction, where $\size{y} < \size{x}$.
This highlights another counter-intuitve property of $\AFL$: 
a system can never outpace the ideal system, as both take the same amount of time to write one target token.
This internal symmetry between the ideal and the actual leads to a potentially undesirably external asymmetry between falling behind and getting ahead.
Deterministic systems, such as wait-$k$, may estimate an emission rate based on summary statistics such as average source and target lengths.
These fixed emission rates will sometimes under-estimate the rate of the ideal system, and they will fall behind and have increased latency (Table~\ref{tab:exAFL_UnequalLengths}, left).
However, they will also sometimes over-estimate the emission rate of the ideal system (Table~\ref{tab:exAFL_UnequalLengths_2}).
In these cases, $\AFL$ will not reward them for outpacing the ideal, forcing them to slow down and receive a lag based on any initial delays.
%
%
It is debatable whether this external asymmetry is worth the internal symmetry and the (arguably desirable) property of avoiding having a wait-1 systems receive a lag of 0, as occurs in Table~\ref{tab:exAFL_UnequalLengths_2}.
If this becomes important, one can sidestep the issue by using a fixed $d$ for $\AFL$'s $g'_d$ rather than defining it as $d=\size{x}/\size{y}$.

\subsection{An empirical comparison}

\begin{figure}[t]
  \begin{center}
    \includegraphics[width=0.45\textwidth]{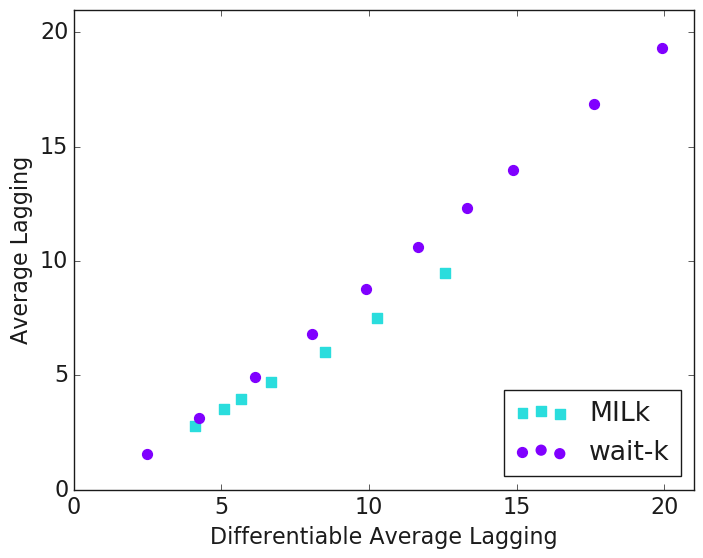}
    \caption{$\AL$ versus $\AFL$ for two types of systems: deterministic (wait-$k$) and adaptive (MILk), as various latency settings for both systems. Results are on the WMT15 German-to-English test set.}
    \label{fig:DALvsAL}
  \end{center}
\end{figure}

In a separate effort, we developed and tested an adaptive streaming NMT model called
Monotonic Infinite Lookback Attention, or MILk~\cite{Avari2019}.
It is trained to minimize a joint objective combining $\AFL$ with likelihood, and has a trade-off
parameter that allows it to vary its latency, similar to wait-$k$'s $k$.
In Figure~\ref{fig:DALvsAL}, we measure both $\AL$ and $\AFL$ for both systems at a variety of latency settings
on WMT15 German-to-English test data.
As one can see, there is a predominantly linear relationship between the two metrics,
with $\AFL$ being more conservative and assigning slightly higher lags.
More worrying is that the slope of this linear relationship is not the same for the two systems:
$\AL$ assigns even lower lags to the adaptive MILk system.
This is despite MILk having been explicitly trained to optimize $\AFL$.
Figure~\ref{fig:DALvsAL} suggests that $\AL$ is likely to favor any adaptive system.

\subsection{Summary of properties of $\AFL$}
We have shown a number of examples and datapoints; we will take this space to concisely summarize some properties of $\AFL$:
\begin{itemize}
    \item Both $\AL$ and $\AFL$ assign lags of $k$ to wait-$k$ systems when $\size{y}=\size{x}$, making them both very interpretable.
    \item Both $\AL$ and $\AFL$ penalize systems for falling behind the ideal emission rate. 
    \item Only $\AL$ rewards systems for outpacing the ideal emission rate. 
    \item Time-indexed lags of $\AFL$ are lower-bounded at $(t-1)d$, mirroring the ideal system. Thus $\AFL$ itself is lower-bounded at 0, unlike $\AL$, which can be negative.
    \item $\AFL$ handles antagonistic cases mishandled by $\AL$, through an underlying model where one can never recover from lag once that lag has been incurred.
\end{itemize}

\section{Conclusion}
We have presented a modified version of Average Lagging dubbed
Differentiable Average Lagging.
By beginning with clear assumptions about how long it takes to write
each target token, we have created a metric that is internally consistent
in its treatment of timing, and which is also differentiable.
%

\section*{Acknowledgments}
Thanks to Naveen Arivazhagan, Wolfgang Macherey and Gaurav Kumar for
feedback on earlier versions of this work.

\bibliography{simul_eval}

\begin{thebibliography}{9}
\expandafter\ifx\csname natexlab\endcsname\relax\def\natexlab#1{#1}\fi

\bibitem[{Alinejad et~al.(2018)Alinejad, Siahbani, and Sarkar}]{Alinejad2018}
Ashkan Alinejad, Maryam Siahbani, and Anoop Sarkar. 2018.
\newblock \href {http://aclweb.org/anthology/D18-1337} {Prediction improves
  simultaneous neural machine translation}.
\newblock In \emph{Proceedings of the 2018 Conference on Empirical Methods in
  Natural Language Processing}, pages 3022--3027. Association for Computational
  Linguistics.

\bibitem[{Arivazhagan et~al.(2019)Arivazhagan, Cherry, Macherey, Chiu, Yavuz,
  Pang, Li, and Raffel}]{Avari2019}
Naveen Arivazhagan, Colin Cherry, Wolfgang Macherey, Chung-Cheng Chiu, Semih
  Yavuz, Ruoming Pang, Wei Li, and Colin Raffel. 2019.
\newblock Monotonic infinite lookback attention for simultaneous machine
  translation.
\newblock In \emph{Proceedings of the 57th Annual Meeting of the Association
  for Computational Linguistics (ACL)}.
\newblock (to appear).

\bibitem[{Bangalore et~al.(2012)Bangalore, Rangarajan~Sridhar, Kolan, Golipour,
  and Jimenez}]{Bangalore2012}
Srinivas Bangalore, Vivek~Kumar Rangarajan~Sridhar, Prakash Kolan, Ladan
  Golipour, and Aura Jimenez. 2012.
\newblock \href {http://aclweb.org/anthology/N12-1048} {Real-time incremental
  speech-to-speech translation of dialogs}.
\newblock In \emph{Proceedings of the 2012 Conference of the North American
  Chapter of the Association for Computational Linguistics: Human Language
  Technologies}, pages 437--445. Association for Computational Linguistics.

\bibitem[{Cho and Esipova(2016)}]{Cho16}
Kyunghyun Cho and Masha Esipova. 2016.
\newblock \href {http://arxiv.org/abs/1606.02012} {Can neural machine
  translation do simultaneous translation?}
\newblock \emph{CoRR}, abs/1606.02012.

\bibitem[{Grissom~II et~al.(2014)Grissom~II, He, Boyd-Graber, Morgan, and
  Daum{\'e}~III}]{Grissom2014}
Alvin Grissom~II, He~He, Jordan Boyd-Graber, John Morgan, and Hal
  Daum{\'e}~III. 2014.
\newblock \href {https://doi.org/10.3115/v1/D14-1140} {Don't until the final
  verb wait: Reinforcement learning for simultaneous machine translation}.
\newblock In \emph{Proceedings of the 2014 Conference on Empirical Methods in
  Natural Language Processing (EMNLP)}, pages 1342--1352. Association for
  Computational Linguistics.

\bibitem[{Gu et~al.(2017)Gu, Neubig, Cho, and Li}]{Gu2017}
Jiatao Gu, Graham Neubig, Kyunghyun Cho, and Victor~O.K. Li. 2017.
\newblock \href {http://aclweb.org/anthology/E17-1099} {Learning to translate
  in real-time with neural machine translation}.
\newblock In \emph{Proceedings of the 15th Conference of the European Chapter
  of the Association for Computational Linguistics: Volume 1, Long Papers},
  pages 1053--1062. Association for Computational Linguistics.

\bibitem[{Ma et~al.(2018)Ma, Huang, Xiong, Liu, Zhang, He, Liu, Li, and
  Wang}]{Ma2018}
Mingbo Ma, Liang Huang, Hao Xiong, Kaibo Liu, Chuanqiang Zhang, Zhongjun He,
  Hairong Liu, Xing Li, and Haifeng Wang. 2018.
\newblock \href {http://arxiv.org/abs/1810.08398} {{STACL:} simultaneous
  translation with integrated anticipation and controllable latency}.
\newblock \emph{CoRR}, abs/1810.08398.

\bibitem[{Oda et~al.(2014)Oda, Neubig, Sakti, Toda, and Nakamura}]{Oda2014}
Yusuke Oda, Graham Neubig, Sakriani Sakti, Tomoki Toda, and Satoshi Nakamura.
  2014.
\newblock \href {https://doi.org/10.3115/v1/P14-2090} {Optimizing segmentation
  strategies for simultaneous speech translation}.
\newblock In \emph{Proceedings of the 52nd Annual Meeting of the Association
  for Computational Linguistics (Volume 2: Short Papers)}, pages 551--556.
  Association for Computational Linguistics.

\bibitem[{Rangarajan~Sridhar et~al.(2013)Rangarajan~Sridhar, Chen, Bangalore,
  Ljolje, and Chengalvarayan}]{RangarajanSridhar2013}
Vivek~Kumar Rangarajan~Sridhar, John Chen, Srinivas Bangalore, Andrej Ljolje,
  and Rathinavelu Chengalvarayan. 2013.
\newblock \href {http://aclweb.org/anthology/N13-1023} {Segmentation strategies
  for streaming speech translation}.
\newblock In \emph{Proceedings of the 2013 Conference of the North American
  Chapter of the Association for Computational Linguistics: Human Language
  Technologies}, pages 230--238. Association for Computational Linguistics.

\end{thebibliography}
\bibliographystyle{acl_natbib}

\end{document}